# A Role-specific Guided Large Language Model for Ophthalmic Consultation Based on Stylistic Differentiation

Laiyi Fu, Binbin Fan, Hongkai Du, Yanxiang Feng, Chunhua Li, Huping Song

***Abstract*—** Ophthalmology consultations are crucial for diagnosing, treating, and preventing eye diseases. However, the growing demand for consultations exceeds the availability of ophthalmologists. By leveraging large pre-trained language models, we can design effective dialogues for specific scenarios, aiding in consultations. Traditional fine-tuning strategies for question-answering tasks are impractical due to increasing model size and often ignoring patient-doctor role function during consultations. In this paper, we propose EyeDoctor, an ophthalmic medical questioning large language model that enhances accuracy through doctor-patient role perception guided and an augmented knowledge base with external disease information. Experimental results show EyeDoctor achieves higher question-answering precision in ophthalmology consultations. Notably, EyeDoctor demonstrated a 7.25% improvement in Rouge-1 scores and a 10.16% improvement in F1 scores on multi-round datasets compared to second best model ChatGPT, highlighting the importance of doctor-patient role differentiation and dynamic knowledge base expansion for intelligent medical consultations. EyeDoc also serves as a free available web based service and souce code is available at https://github.com/sperfu/EyeDoc.

## I. Introduction

OPHTHALMIC consultations are a critical component of the medical field [21], involving the diagnosis, treatment and prevention of eye diseases. With an aging population and the impact of modern lifestyles, eye health issues are becoming increasingly prevalent, driving up the demand for ophthalmology consultations. Doctors typically diagnose eye diseases by examining symptoms, physical signs, and imaging test results. Recently, the demand for online consultations has surged due to the growth of internet and telemedicine technologies. People are increasingly opting for online consultations to communicate with doctors, which saves the time and effort of visiting hospitals and waiting in long queues. This trend is particularly beneficial for those living in remote areas with limited access to specialized healthcare services [30,31]. Online consultations save time and effort, especially for those in remote areas with limited access to specialized healthcare. They offer convenience, flexibility, and reduced travel costs. The COVID-19 pandemic has further accelerated the adoption of telemedicine [22,28], making it crucial for maintaining healthcare services while minimizing infection risk.

However, there are some pain points in ophthalmology consultation. Firstly, the number of ophthalmologists in some areas is insufficient, leading to long waiting times and negatively impacting patient experience. Secondly, some primary care doctors lack the expertise to accurately diagnose and treat ophthalmic conditions, resulting in a high rate of misdiagnosis [31]. Lastly, the traditional ophthalmology consultation process often requires patients to visit the hospital in person, increasing their time and cost burden while also limiting the efficiency of doctors' diagnoses and treatments. These issues highlight the need for innovative solutions to improve the accessibility and quality of ophthalmic care, both in-person and online.

In recent years, with the continuous progress of Natural Language Processing (NLP) technology and Large Language Models (LLMs), LLMs have been gradually applied to various vertical domains within the healthcare industry due to their powerful semantic understanding and knowledge generalization capabilities. In the biomedical crossover, Jinhyuk Lee proposed BioBERT [5], a biomedical domain synthesis model pre-trained on a large biomedical corpus, which achieved commendable performance in several biomedical natural language processing tasks such as bio-knowledge quizzing. What's more, In the field of intelligent medical consultation, Liu et al. collected a large dataset of entity-labeled medical consultation dialogues and constructed an entity-aware medical consultation algorithm based on MedDG [7]. The ChatDoctor [8] algorithm uses the collected medical document information as an external knowledge source for assisted medical diagnosis in the form of autonomous knowledge retrieval. In the Chinese medical consultation environment, BenTsao [4] generates and processes data for Chinese medical knowledge graph with the semantic

Binbin Fan, Hongkai Du and Yanxiang Feng are with the School of Automation Science and Engineering, Xi'an Jiaotong University, Xi'an, Shannxi, 710049, China. Chunhua Li and Huping Song are with Ophthalmology Department, Xi'an People's Hospital (Xi'an Fourth Hospital), Xi'an, Shannxi, 710049, China.

Laiyi Fu is with the School of Automation Science and Engineering, Xi'an Jiaotong University, Xi'an, Shannxi, 710049; Research Institute of Xi'an Jiaotong University, Zhejiang, 311200; Sichuan Digital Economy Industry Development Research Institute, 610036, China.
Correspondence should be addressed to L.F. & Huping Song(laiyifu@xjtu.edu.cn & songhpxian@163.com)
Conflict of Interest: none declared.

understanding capability of ChatGPT [2], and completes the migration of the generalized big model to the medical consultation domain based on the efficient parameter fine-tuning strategy of LoRA [12]. Shenzhen Big Data Research Institute organizes text data in the medical field and obtains HuatuoGPT [1] medical consultation big model through supervised learning and hybrid reinforcement learning. Wang et al. [28] comprehensively analyzed and enhanced the E4 communication model established in online medical consultation in China by proposing a new E5 model. In addition to pure text data, ConVIRT [9] has launched a multimodal exploration in the medical consultation domain, which combines medical images with the consultation process to provide patients with comprehensive diagnostic conclusions.

However, current techniques for constructing and applying large medical Q&A models still face the following several challenges [33]: (1) Specialization and Complexity: The medical field's high degree of specialization and complexity demands that Q&A models possess extensive medical knowledge and advanced processing capabilities. Most existing medical Q&A models are designed for general medical topics and lack the specificity required for particular specialties like ophthalmology. This limitation prevents the models from providing accurate and professional responses in specialized areas. (2) Role and Style Differentiation: Current medical consultation models often fail to differentiate between the roles and communication styles of doctors and patients. This shortcoming affects the models' ability to accurately address the core needs of both parties during consultations[24,29]. Doctor-patient communication has distinct characteristics[6]: patients primarily describe and express their symptoms, while doctors focus on medical terminology, diagnostic criteria, and personalized treatment plans [34]. (3) Data Scarcity and Diversity: The medical field suffers from data scarcity and diversity [6,7]. The limited availability of training data for rare or emerging diseases hinders the models' ability to recognize and address these conditions effectively [3]; Additionally, significant variations in medical data across different languages and cultures pose challenges in constructing cross-lingual and cross-cultural medical consultation models.

In order to address the aforementioned challenges, this paper proposes a novel ophthalmic medical questioning algorithm (EyeDoctor/EyeDoc) based on large language model. This algorithm leverages doctor-patient role perception and an augmented knowledge base with exogenous disease information, aiming to improve the model's question-and-answer accuracy while enhancing its ability to deal with disease counseling in new contexts. We evaluated the model using multiple real-world datasets and question-and-answer scenarios. The experimental results demonstrate that our framework achieves effective and satisfactory performance in ophthalmology question-and-answer contexts. Additionally, a series of ablation experiments highlighted the importance of considering doctor-patient role characteristics and dynamically expanding the knowledge base for intelligent medical consultations. To further validate and optimize our model in practical applications, we have developed an online ophthalmology consultation platform, which is available to users for free. The contributions of our work are as follows:

- We gather over one million single-round and 10,000 multi-round eye consultation dialogues from the Quick Ask Doctor and Ding Xiang Doctor websites. Utilizing ChatGPT [2]'s advanced semantic understanding, we de-identified and standardized this data creating robust datasets for the EyeDoctor algorithm and future research.
- We propose a role-specific large language model framework with a knowledge base of 519 ophthalmic diseases. Integrating this through similarity matching algorithm, we enriched dialogues, reducing response unpredictability and enhancing the model proficiency. This allows for continuous dataset updates without retraining, improving reusability and adaptability.
- We introduced the EyeDoctor framework, which uses a multi-module pre-training model to differentiate dialogue roles. It separately extracts linguistic styles and semantics of doctors and patients, delivering high-precision responses for both single-round and multi-round consultations through knowledge enhancement and role-differentiated fine-tuning.

## II. DATA PROCESSING

Current research in the field of medical consultation mainly focuses on data-driven domain adaptation of generalized macromodels through simulated data generated from structured medical documents or medical knowledge graphs. Given the scarcity of real conversation data for ophthalmic disease diagnosis, this paper systematically collect different types of data from three Chinese online resources: QuickAskDoctor's website which is an online medical and health service platform created by Zhuhai Health Cloud Technology Co., Ltd; Ding Xiang Doctor's website which is a medical and healthcare service platform with legal Internet hospital license developed by Lilac Garden team, and 99 Health Net which is an online medical health platform, dedicated to providing users with comprehensive and detailed disease information and medical service resources. Details on the dataset and prompt design are in Supplementary Materials.

We design prompt statements [10] for the ophthalmic consultation dialogues data by using the powerful semantic understanding and text generation capabilities of ChatGPT 3.5 turbo API. This preprocessing step aims to remove irrelevant and potentially sensitive information, standardize medical terminology, and ensure that the data aligns with clinical realities in ophthalmology. This provides a solid data foundation for the construction of efficient and accurate ophthalmology intelligent consultation algorithms.

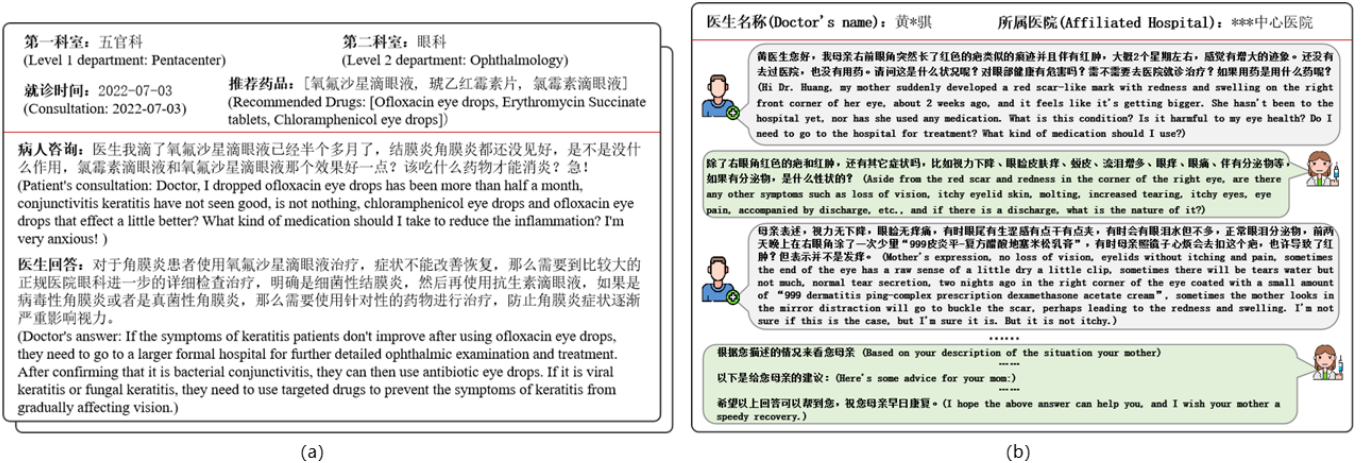

Fig. 1. (a) This is an example of a single round of questioning; (b) This is an example of multiple rounds of questioning.

In terms of single-turn data, we focus on the medical Q&A data of QuickAskDoctor's website during the period of June 1 to November 10, 2022, and collect 1,292,012 single rounds of data from all departments, covering 19 first-level departments such as pediatrics, internal medicine, surgery, and 127 second-level departments. In addition to the single-round conversation data, additional information such as the type of department visited, time of visit, and recommended medications were also collected during data collection. Then, we will use these single-round data to perform unsupervised fine-tuning of the RoBERT, a large language model to facilitate further subsequent characterization of the doctor-patient relationship and external knowledge sources.

In terms of multi-turn data,, we collect ophthalmology medical consultations from the website of Ding Xiang Doctor, which contains 8,647 multi-round conversations from 108 doctors. The average number of conversation rounds for a single consultation is 6.27, and the average number of conversations for a single doctor is 80.06. And the raw data needs to be further processed with sensitive information such as the doctor's name and the standardization of medical terminology. Eventually, the processed conversation data will be applied to the following experiments to construct intelligent consultation algorithms for real ophthalmology environments.

What's more, in order to improve the algorithm's ability to deal with emerging and rare diseases in the process of ophthalmic disease diagnosis [8,27], we collect and organize 519 kinds of ophthalmic disease knowledge information from the 99 Disease Database website. Based on the disease knowledge information, we construct a document knowledge base, which is convenient for dynamic knowledge expansion of the algorithm. When constructing the ophthalmic disease knowledge base, 10 kinds of disease information including disease name, overview, symptoms, treatment plan, examination, identification, etiology, prevention, complications, and disease medication are collected in order.

III. METHOD

*A. Algorithmic Framework for Intelligent Medical Questioning*

In EyeDoctor, a patient u initiates a dialogue by providing the model with information such as disease symptoms. The entire dialogue can be represented as $D = \{S_t^u, S_t^u\}_{t=1}^T$, where T represents the number of dialogue turns. A dialogue turn t consists of the user's dialogue request $S_t^u = \{w_n\}_{n=1}^N$ and the algorithm's response $S_t^r = \{w_n\}_{n=1}^N$, where $w$ represents a word from the vocabulary $W$, $N$ represents the length of the dialog, n represents the nth word in the dialog statement. During the dialogue, the disease document knowledge base $k \in K$ serves as an external knowledge source to augment the preceding dialogue, enhancing the model's diagnostic capabilities [35]. Here, $K = \{w_n\}_{n=1}^N$ represents a single piece of disease information.

*B. Knowledge Representation Model for Medical Consultations*

In the highly specialized domain of medicine, there are a large number of medical terminologies and unique text organization and information expression patterns, and it is difficult for a generic pre-trained language model that has not been customized and optimized to fully demonstrate excellent expressiveness and adaptability [23]. For this reason, based on a large amount of medical text single-round data collected in this paper, the original RoBERTa model is fine-tuned to the medical domain using word masking (MLM) [6], and a specialized model (DiagBERT) [15] for the medical consultation domain is constructed to achieve more accurate and efficient processing and understanding of complex text information in this domain. The constructed formulas are as follows:

$$D^{'} = Masked(D). \qquad (1)$$

$$CLS_{D^{'}}, E_{x_1:x_n} = RoBERTa(Embedding(D^{'})). \qquad (2)$$

Where $D$ denotes the content of a single round of conversation, $D'$ denotes the content of the conversation initialized with a random word mask and input, $CLS_{D'}$ denotes the [CLS] labels, and $E_{x_1:x_n}$ denotes the hidden state vector for each word of $D'$.

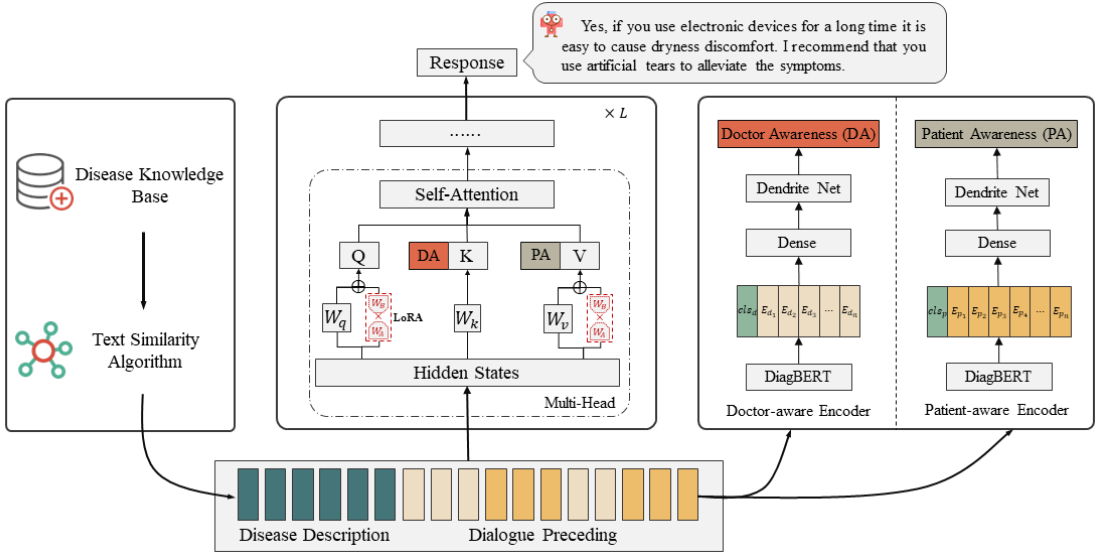

Fig. 2. EyeDoctor model scheme. EyeDoctor enhances ophthalmology consultation accuracy through doctor-patient role perception and an augmented knowledge base. The model integrates a Disease Knowledge Base, a Text Similarity Algorithm, and a multi-head Self-Attention mechanism. The Self-Attention incorporates Doctor Awareness (DA) and Patient Awareness (PA) modules via DiagBERT encoders, utilizing dendritic and dense networks to process role-specific information, leading to precise response generation.

For the masked word $w_i^*$, its corresponding hidden state vector $E_{x_i^*}$ is taken and the predicted probability distribution of all the words on the vocabulary is generated by a linear layer and softmax function with the following formula:

$$P(w_i^*|D) = softmax(WE_{x_i^*} + b). \quad (3)$$

Where $W$ and $b$ are the training parameters.

The word mask training approach performs parameter optimization by calculating the cross-entropy loss between the predicted probabilities and the true labels with a loss function expressed as follows:

$$L_{MLM} = \sum_{i \in mask\ position} -log\ P(w_i^*|D). \quad (4)$$

After unsupervised training, a knowledge representation model DiagBERT [15] dedicated to the medical consultation domain can be obtained. In the framework of intelligent consultation algorithms DiagBERT will be applied in the tasks of disease knowledge representation and doctor-patient role representation.

### C. Doctor-Patient Role Dual Coding Learner

Most of the existing medical consultation algorithms rely on massive medical consultation data to optimize diagnostic accuracy and treatment plans, and although these models perform well in general disease diagnosis and treatment, they largely ignore the significant group differences between doctors and patients [29]. In the real medical consultation scenario, the patient, as the person who experienced the disease, provides information about the specific manifestations of the disease, the duration of the cycle, and other symptom characteristics [32]. Doctors, as providers of medical services, have professional knowledge background and clinical practice experience, and they pay more attention to customizing personalized diagnosis and treatment plans and preventive measures according to the patient's condition during the consultation and communication. Therefore, this paper proposes a dual coding learner for doctor-patient roles to more accurately simulate the real doctor-patient consultation scenarios and improve the algorithm performance and user experience.

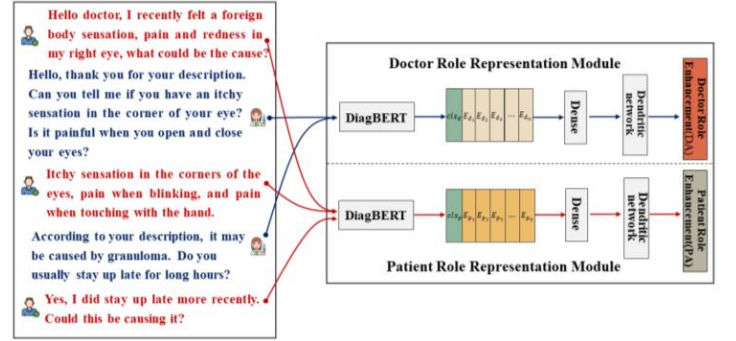

Fig. 3. Doctor-Patient Role Dual Coding Learner.

The doctor-patient role dual coding learner is shown in Fig.7. The dialog above statements are first divided according to the doctor role and the patient role. The conversation history of the doctor at turn t is denoted as $D_d = \{s_m^r\}_{m=1}^{t-1}$ and the conversation history of the patient is denoted as $D_p = \{s_m^r\}_{m=1}^{t-1}$. Both of them splice the conversation history and go through the DiagBERT model [15] respectively to obtain the high-dimensional semantic characterization, which is computed as follows:

$$CLS_d, H_d = DiagBERT(Embedding(D_d)). \quad (5)$$

$$CLS_p, H_p = DiagBERT(Embedding(D_p)). \quad (6)$$

Where $H_d$, $H_p$ represent the high-dimensional semantic representations of the doctor and patient conversation above, respectively, and $CLS_d$, $CLS_p$ denote the unified representations of the doctor and patient input texts.

After obtaining the unified representations of the doctor and the patient, both will learn their respective linguistic features based on a two-layer dendritic network, respectively, and there is no shared weight between the doctor and the patient representations.

$$T_p = Dense(CLS_p). \tag{7}$$

$$H'_p = W_{P_2}((W_{P_1}T_p \otimes T_p) \otimes T_p. \tag{8}$$

$$T_d = Dense(CLS_d). \tag{9}$$

$$H'_d = W_{d_2}((W_{d_1}T_d) \otimes T_d. \tag{10}$$

Where $W_{P_1}, W_{P_2}, W_{d_1}, W_{d_2}$ are the training parameters of the dendritic network, $\otimes$ is the Hadamard Product, and $H'_p$ and $H'_d$ are the final physician role augmentation representation and patient role augmentation representation. Unlike the traditional form of neural network $Wx + b$, the dendritic network simulates the biological dendritic process and employs the form of elemental product to effectively establish effective logical interactions between the input $CLS_x$ and the results of each layer.

### D. Knowledge Enhancement Algorithm Based on Disease Knowledge Repository

In today's fast-changing medical field, the emergence of emerging and rare diseases and the continuous advancement of disease diagnosis and treatment methods pose a serious challenge to the existing knowledge-consolidated medical diagnosis algorithms. Emerging and rare diseases often have unique clinical manifestations, diagnostic criteria, and treatment strategies, which require highly specialized theoretical knowledge and the ability to acquire the latest medical information. Therefore, in the design of medical consultation algorithms, a knowledge enhancement algorithm based on the disease knowledge base is proposed to ensure the scalability of the diagnostic logic and the implementation of the treatment recommendations of the consultation model, in order to enhance the adaptability of the medical consultation model in the face of emerging and rare diseases [8].

Among the collected knowledge information of a single disease, the disease overview concisely summarizes the main features of the current disease; the disease symptoms list in detail the various signs and discomforts that the patient may exhibit, providing strong support for the diagnostic algorithm to analyze the details of the symptoms mentioned by the user; and the disease treatment protocols cover the mainstream treatment protocols in the current medical practice for the diagnostic model to refer to these treatment protocols when generating responses, ensuring the professionalism and accuracy of the recommendations. These three pieces of information are crucial for the construction of the disease knowledge enhancement algorithm. Therefore, the knowledge information of a single disease document is constructed in the order of disease name, disease overview, disease symptoms, disease treatment, and other information.

After constructing the disease knowledge base, the DiagBERT disease knowledge characterization model is used to accurately characterize the knowledge of each disease knowledge document in the conversation above $D$ and the disease knowledge base $K$ using the DiagBERT disease knowledge characterization model. The knowledge characterization algorithm for the conversation above D and a single disease document knowledge K is shown below:

$$CLS_D, H_D = DiagBERT(Embedding(D)). \tag{11}$$

$$CLS_K, H_K = DiagBERT(Embedding(K)). \tag{12}$$

Where $CLS_D$, $CLS_K$ denote the textual semantic representation of the dialog above content D and disease knowledge K.

Next, the semantic similarity between the current dialog above and all diseases in the disease knowledge base K is calculated using the cosine similarity matching algorithm. For the dialog above D and a single disease document knowledge K, the similarity calculation formula is as follows:

$$sim_{D,K} = \frac{CLS_D \cdot CLS_K}{\|CLS_D\| \|CLS_K\|}. \tag{13}$$

After calculating the similarity $sim_{D,K}$ between the dialog D above and each disease document in the knowledge base K, the disease knowledge document $K_{sim}$ with the highest similarity is selected as an external knowledge source for knowledge augmentation of the dialog content.

### E. Parameter tuning strategy

In order to solve the problems encountered in the process of full parameter fine-tuning of large language models in terms of data, arithmetic power, compatibility, deployment, and many other aspects, we decided to adopt the optimization strategy of large language models with efficient parameter fine-tuning as the core. In view of the diversity of efficient parameter fine-tuning and the difficulty of selection, this paper adopts an efficient parameter fine-tuning method that integrates Prefix Tuning [11] and LoRA [12] to fine-tune the parameters of e.g., LLaMA [17,18] big language model, and constructs a medical consultation algorithm for doctor-patient role perception and disease knowledge enhancement.

We combine the low-rank matrices $W_{dowm} \in \mathbb{R}^{d \times r}$ and $W_{up} \in \mathbb{R}^{d \times r}$ to be trained with the Query and Value projection matrices in multi-head attention to influence the parameter update process. Assuming that the input to the multi-head attention in layer l is $h_l \in \mathbb{R}^{d \times r}$, the fusion formula for Query is shown below:

$$Q = (W_Q^T + \alpha \cdot W_{dowm}{}^T W_{up}{}^T)h_l. \tag{14}$$

Where $W_Q^T \in \mathbb{R}^{d \times d}$ is the original weight matrix with parameters frozen during training and α is the hyperparameter.

On the other hand, this paper employs the Prefix Tuning [11]

parameter fine-tuning strategy to integrate the doctor role enhancement $H_p^{'}$ and patient role enhancement $H_d^{'}$ obtained from the doctor-patient role dual encoding learner into the multi-head attention mechanism. Specifically, $H_p^{'}$ and $H_d^{'}$ are used as the front-attachment vectors for the Key and value parts

$$head_i = Attn(Q, concat(H_d^{'},K), concat(H_p^{'},V)). \quad (15)$$

of the multi-head attention mechanism, respectively, to further guide the dialog response generation process. This improved formulation of the multi-head attention mechanism is shown below:

Where $Q$ and $V$ are the results after fine-tuning by LoRA parameters, and K is the original hidden state after linear mapping.

During model training, the model input to the algorithm consists of the highest similarity disease knowledge document $K_{sim}$ and the dialog above $D$ spliced together. The text input is denoted as $X_{input} = \{K_{sim}, D\}$. The training parameters in LoRA [12] are denoted as $\Theta_{lora}$, and the training parameters in Prefix Tuning [11] and Doctor-Patient Role Dual Coding Learner are denoted as $\Theta_{pt}$. The loss function is shown below:

$$L(\Theta_{lora}, \Theta_{pt}) = -\frac{1}{S}\sum_{s=1}^{S} \log Pr(T^{(s)}|X_{input}^{(s)}; \Theta_{lora}, \Theta_{pt}). \quad (16)$$

The entire loss function can be viewed as a parameter optimization process based on great likelihood estimation, where S denotes the number of individual parameter update rounds to the training set. Each piece of data in the dataset contains a textual input $X_{input}^{(s)}$ and a target response $T^{(s)}$. The parameter optimization process can be viewed as a probability distribution that predicts the target response $T^{(s)}$ based on the input $X_{input}^{(s)}$.

Overall, the EyeDoctor model contains a total of four parameters in the RoBERTa model parameter $\Theta_{RoBERTa}$, the baseline bigram model parameter $\Theta_{llm}$, and $\Theta_{lora}$ and $\Theta_{pt}$ in the hybrid efficient parameter fine-tuning strategy. $\Theta_{RoBERTa}$ undergoes the full parameter fine-tuning for healthcare domain adaptation with the loss function of (4). The baseline macromodel parameter $\Theta_{llm}$ is fixed, and $\Theta_{lora}$ and $\Theta_{pt}$ are incorporated into the baseline model via a hybrid efficient parameter strategy with a loss function of (16).

## IV. EXPERIMENT

### A. Evaluation Metrics

In this experiment, we validate the performance of EyeDoctor using several metrics. For the dialog subtask, we use Distinct [13] and Bleu [14] to automate the verification of the model's dialog response diversity and consistency, respectively. $Distinct@n$ calculates the ratio of the number of distinct n-grams appearing in all the text generated content to the total number of words, and a higher score on $Distinct@n$ means that the generated text contains more unique n-grams, i.e., it exhibits higher diversity and the generated text content is richer. $Bleu@n$ calculates the degree of matching between the generated text content and the original text content at the n-gram level, and a higher $Bleu@n$ score means that the generated text is more compliant with the original text. In addition, we also introduce the $ROUGE-l$ evaluation metric to verify the consistency characteristics of the model-generated text.

In order to demonstrate EyeDoctor's medical consultation expertise and semantic accuracy in medical consultation scenarios, we calculates the BERTScore semantic similarity scores for each benchmark model based on the BERT-Base-Chinese knowledge representation model. Compared to traditional evaluation metrics based on rule matching, BERTScore is not constrained by fixed-length n-gram matching rules. It considers global semantic consistency and long-distance dependencies when evaluating texts, achieving fine-grained semantic understanding and providing evaluation results that are closer to human perception. BERTScore evaluates the quality of text generation by calculating the recall $R_{BERT}$, precision $P_{BERT}$, and F1 value $F_{BERT}$ [16].

### B. Baseline Models

In order to practically evaluate the actual effectiveness of EyeDoctor in medical consultation scenarios, two types of methods, namely, the medical professional consultation grand model and the generalized grand language model, were selected as baseline controls in this study. The statistical information of the baseline models is shown in Table I.

TABLE I
INFORMATION STATISTICS FOR THE BASELINE MODEL

| Type | Model Name | Year | Publisher | Number of parameters |
|---|---|---|---|---|
| Medical | HuatuoGPT | 2023 | EMNLP | 7B |
|  | BenTsao | 2023 | - | 7B |
|  | DoctorGLM | 2023 | - | 6B |
| Universal | TinyLLaMA | 2023 | Meta | 1B |
|  | LLaMA-2-7B | 2023 | Meta | 7B |
|  | ChatGLM | 2022 | ACL | 6B |
|  | ChatGPT 3.5 | 2023 | OpenAI | - |

The 3 most recent large models of medical specialty consultations are as follows：

**1)HuatuoGPT[1]:** HuatuoGPT is a Chinese-English medical big language model fine-tuned based on the LLaMA [17,18] general big language model released by Shenzhen Big Data Research Institute. It synthesizes the data extracted from ChatGPT [2] and doctors' real conversation questioning data, and realizes the fusion of the two kinds of data through reinforcement learning approach.

**2)BenTsao[4]:** BenTsao is a knowledge-enhanced Chinese medical big language model proposed by Harbin Institute of Technology. It utilizes the text generation capability of ChatGPT [2] to generate more than 8,000 simulated data for training using the Chinese Medical Knowledge Graph (CMeKG) in the absence of real-world data medical domain data.

**3)DoctorGLM[19]:** DoctorGLM is a large language model focusing on the Chinese medical domain based on the ChatGLM generalized large language model trained using

TABLE II
RESULTS OF AUTOMATED EVALUATION OF SINGLE-ROUND QUESTIONING DATASETS (T-TEST WITH P-VALUE<0.05)

| Type | Name | R-l | Bleu-1 | Bleu-2 | Bleu-3 | Bleu-4 | Dist-1 | Dist-2 | Dist-3 | Dist-4 |
|---|---|---|---|---|---|---|---|---|---|---|
| Medical | HuatuoGPT-7B | 0.2378 | 0.3065 | 0.1358 | 0.0864 | 0.0571 | 0.3720 | 0.7109 | 0.8034 | 0.8632 |
|  | BenTsao-7B | 0.2115 | 0.3824 | 0.1529 | 0.0823 | 0.0569 | 0.4574 | 0.7216 | 0.8349 | 0.8140 |
|  | DoctorGLM-6B | 0.2321 | 0.4003 | 0.1629 | 0.1025 | 0.0669 | 0.4284 | 0.6819 | 0.7416 | 0.7807 |
| Universal | TinyLLaMA | 0.2250 | 0.4483 | 0.2295 | 0.1643 | 0.1262 | 0.4652 | 0.5857 | 0.6123 | 0.6301 |
|  | LLaMA-2-7B | 0.2650 | 0.5067 | 0.2442 | 0.1675 | 0.1213 | **0.5251** | 0.7306 | 0.7764 | 0.8056 |
|  | ChatGLM-6B | 0.2260 | 0.2793 | 0.1249 | 0.0823 | 0.0564 | 0.3262 | 0.6296 | 0.7245 | 0.7871 |
|  | ChatGPT | 0.2622 | 0.3838 | 0.1733 | 0.1147 | 0.0774 | 0.4246 | 0.7327 | 0.8141 | 0.8677 |
| Ours | EyeDoctor-1B | 0.3313 | 0.4838 | 0.2536 | 0.1936 | 0.1542 | 0.4805 | 0.7619 | 0.8169 | 0.8486 |
|  | EyeDoctor-7B | **0.3375** | **0.5305** | **0.2754** | **0.2082** | **0.1638** | 0.5153 | **0.7910** | **0.8471** | **0.8797** |

LoRA efficient parameter fine-tuning strategy. It aims to process and understand Chinese medical dialog scenarios and provide intelligent consultation services.

The 3 widely used generalized large language models are listed below:

**1)LLaMA[17,18]:** LLaMA is a series of large language models developed by Meta, which has been used as an important large modeling base in various verticals by showing comparable or even better performance than larger models through efficient training with a relatively small number of parameters.

**2)ChatGLM[20]:** ChatGLM is a Chinese-English bilingual dialog bot developed by Smart Spectrum AI. It is equipped with multi-domain knowledge reasoning and application capabilities, supports interaction with users through natural language dialogues, handles a variety of natural language tasks, and is now used as a large model base for many Chinese Q&A domains.

**3)ChatGPT [2]:** ChatGPT is an advanced AI dialog model developed by OpenAI research. The model is massively pre-trained on massive datasets and is capable of engaging in fluent, logical conversations with users and performing diverse and complex linguistic tasks such as writing and common-sense reasoning.

### C. Hyperparameterization

EyeDoctor used a NVIDIA A40 graphics card for hardware selection for model parameter optimization. Two parametric versions of the model, EyeDoctor-1B and EyeDoctor-7B, were obtained using TinyLLaMA(1B) and LLaMA-2-7B as the base model, respectively. During the training process, the base model parameters are frozen and are not involved in the model parameter optimization process. RoBERTa-base-Chinese is used as the knowledge fusion learning model for characterizing the disease knowledge base and doctor-patient roles. The maximum length of exogenous knowledge of disease documents was 512 characters, while the maximum length for multi-round conversation history was set to 1024 characters. The rank of the low-rank matrix in LoRA was set to 8, the length of Prefix Tuning was set to 100, the batch size was set to 32, the learning rate was set to 0.0005, and the first 10% of the training set was pre-warmed with a linear learning rate, and the parameter optimization was performed using AdamW. We will train 5 rounds and save the model weights that perform optimally on the validation set. The data set is divided according to the 8:1:1 division of training set, validation set, and test set. For the baseline model, in order to ensure the fairness of the experiment and to play the optimal performance of the baseline model, the same test set is used, and the response text is generated through the form of API calls. Finally, when conducting the experimental investigation, all the measurement indexes were tested several times to ensure that the experimental results meet the demand of statistical significance.

### D. Automated Quantitative Assessment

Tables II and III present the automatic quantization results of different dialog algorithms on the single-round and multi-round questioning datasets collected in this paper. The compared methods are categorized into three groups: healthcare-specific large models, generalized large language models, and the models proposed in this chapter. Both the healthcare-specific large models and the generalized large language model are based on the automated evaluation scheme of previous work, utilizing a zero-shot approach to test the experimental effects on the test set [1].

From the automated evaluation results, among the medical-specific baseline models, HuatuoGPT-7B outperforms other medical models in overall performance due to its extensive knowledge base of Chinese medical literature and numerous medical single-round Q&A datasets. DoctorGLM-6B contains a large number of repetitive statements in the generated medical responses, resulting in overall performance inferior to BenTsao-7B medical questioning model. In terms of general-purpose large language models, ChatGPT [2], as a popular high-performance large language model today, has achieved better results in general-purpose natural language processing domain tests.

The LLaMA-2-7B benchmark model chosen in this paper is fully parameter fine-tuned based on massive Chinese dataset, which has obvious advantages in processing Chinese data. Therefore, ChatGPT [2] and Llama-2-7B models have close performance, both surpassing TinyLLaMA and ChatGLM-6B. In the automated evaluation validation, the proposed model EyeDoctor outperforms all of the above benchmark models, with a particularly notable advantage in the multi-round questioning dataset. For instance, Rouge-l scores are ahead of ChatGPT [2] by 7.25% on the single-round dataset and 10.16% on the multi-round dataset. This demonstrates both the professional performance of the models mentioned in this paper in specialized medical consultations and the importance of collecting multi-round conversation datasets in the field of medical consultations.

TABLE III
RESULTS OF AUTOMATED ASSESSMENT OF MULTI-ROUND QUESTIONING DATASETS (T-TEST WITH P-VALUE<0.05)

| Type | Name | R-l | Bleu-1 | Bleu-2 | Bleu-3 | Bleu-4 | Dist-1 | Dist-2 | Dist-3 | Dist-4 |
|---|---|---|---|---|---|---|---|---|---|---|
| Medical | HuatuoGPT-7B | 0.2631 | 0.2773 | 0.1580 | 0.1137 | 0.0840 | 0.4844 | 0.7872 | 0.8567 | 0.8991 |
| | BenTsao-7B | 0.2342 | 0.2553 | 0.1557 | 0.0838 | 0.0518 | 0.4464 | 0.7132 | 0.8184 | 0.8240 |
| | DoctorGLM-6B | 0.1967 | 0.2454 | 0.0962 | 0.0517 | 0.0301 | 0.4678 | 0.7463 | 0.8082 | 0.8472 |
| Universal | TinyLLaMA | 0.2598 | 0.3317 | 0.1532 | 0.0937 | 0.0629 | 0.5948 | 0.7756 | 0.8144 | 0.8410 |
| | LLaMA-2-7B | 0.2706 | 0.3448 | 0.1634 | 0.1016 | 0.0678 | 0.6189 | 0.8198 | 0.8562 | 0.8786 |
| | ChatGLM-6B | 0.2605 | 0.2981 | 0.1631 | 0.1160 | 0.0853 | 0.5000 | 0.7832 | 0.8467 | 0.8843 |
| | ChatGPT | 0.2762 | 0.2962 | 0.1667 | 0.1194 | 0.0878 | 0.5164 | 0.8172 | 0.8795 | 0.9180 |
| Ours | EyeDoctor-1B | 0.3617 | **0.4841** | 0.2797 | 0.1987 | 0.1501 | **0.6846** | 0.8885 | 0.9123 | 0.9250 |
| | EyeDoctor-7B | **0.3778** | 0.4829 | **0.2908** | **0.2123** | **0.1637** | 0.6707 | **0.8886** | **0.9149** | **0.9295** |

In addition, compared to the the smaller parameter EyeDoctor-1B model, EyeDoctor-7B with nearly seven times more parameters, shows superior performance. All indicators for EyeDoctor-7B surpass those of EyeDoctor-1B on the single-round dataset, and seven indicators are ahead on the multi-round dataset, underscoring the performance enhancements brought by the expansion of the model's parameter scale.

### E. Evaluating the Semantic Accuracy of Medical Questioning

To show the semantic accuracy of EyeDoctor model for medical consultation, this chapter calculates BERTScore semantic similarity score based on BERT-Base-Chinese knowledge representation model. The experimental results are shown in Tables IV and V.

TABLE IV
RESULTS OF SEMANTIC ACCURACY EVALUATION OF SINGLE-ROUND QUESTIONING DATASET (T-TEST WITH P-VALUE<0.05)

| Type | Name | Recall | Precision | F1 |
|---|---|---|---|---|
| Medical | HuatuoGPT-7B | 0.6925 | 0.6242 | 0.6563 |
| | BenTsao-7B | 0.5834 | 0.6703 | 0.6229 |
| | DoctorGLM-6B | 0.6451 | 0.6428 | 0.6432 |
| Universal | TinyLLaMA | 0.5919 | 0.6358 | 0.6119 |
| | LLaMA-2-7B | 0.6385 | 0.6656 | 0.6511 |
| | ChatGLM-6B | 0.6767 | 0.6318 | 0.6531 |
| | ChatGPT | 0.6844 | 0.6571 | 0.6699 |
| Ours | EyeDoctor-1B | **0.6985** | 0.6696 | 0.6829 |
| | EyeDoctor-7B | 0.6947 | **0.6889** | **0.6912** |

From the results of medical consultation accuracy evaluation, in the medical-specific baseline model, HuatuoGPT-7B and BenTsao-7B perform close to each other, and outperform DoctorGLM-6B on the single-round and multi-round medical consultation datasets. Among them, on the single-round dataset, HuatuoGPT-7B outperforms BenTsao-7B on Recall and F1 scores. In the multi-round dataset, BenTsao-7B outperforms HuatuoGPT-7B in Precision and F1 scores. among the generalized big language models, ChatGPT [2] outperforms other big language models by virtue of its massive knowledge accumulation. Compared with all the previous competitors, out proposed model outperforms all the above baseline models, especially the advantage is more obvious in the multi-round questioning dataset. Compared with ChatGPT [2], our model leads 1.03%, 3.18%, and 2.13% in Recall, Precision, and F1 values evaluated on the single-round dataset, and 5.47%, 8.83%, and 7.2% on the multi-round dataset, respectively. In terms of parameter scale, EyeDoctor-7B achieves better performance in single-round and multi-round questioning semantic accuracy evaluation due to the nearly seven times parameter scale of EyeDoctor-1B, and EyeDoctor-7B outperforms the EyeDoctor-1B model in five out of a total of six evaluation metrics.

TABLE V
RESULTS OF SEMANTIC ACCURACY EVALUATION OF SINGLE-ROUND QUESTIONING DATASET (T-TEST WITH P-VALUE<0.05)

| Type | Name | Recall | Precision | F1 |
|---|---|---|---|---|
| Medical | HuatuoGPT-7B | 0.7054 | 0.6397 | 0.6699 |
| | BenTsao-7B | 0.6763 | 0.7211 | 0.6952 |
| | DoctorGLM-6B | 0.6169 | 0.5621 | 0.5872 |
| Universal | TinyLLaMA | 0.6797 | 0.6491 | 0.6630 |
| | LLaMA-2-7B | 0.7010 | 0.6578 | 0.6776 |
| | ChatGLM-6B | 0.6858 | 0.6430 | 0.6626 |
| | ChatGPT | 0.7027 | 0.6458 | 0.6723 |
| Ours | EyeDoctor-1B | 0.7410 | 0.7290 | 0.7337 |
| | EyeDoctor-7B | **0.7574** | **0.7341** | **0.7443** |

### F. Ablation Experiment in Doctor-Patient Role Perception and Knowledge Enhancement

In order to show the influence of exogenous knowledge base and doctor-patient role representation on medical consultation results in single- and multi-round medical consultation scenarios, we designed an ablation experiment in the same experimental environment for experimental validation. In the ablation experiment, "no knowledge base" means removing the exogenous knowledge augmentation module, and "no doctor-patient representation" means removing the doctor-patient role dual-coding learner. The results are shown in Fig.8.

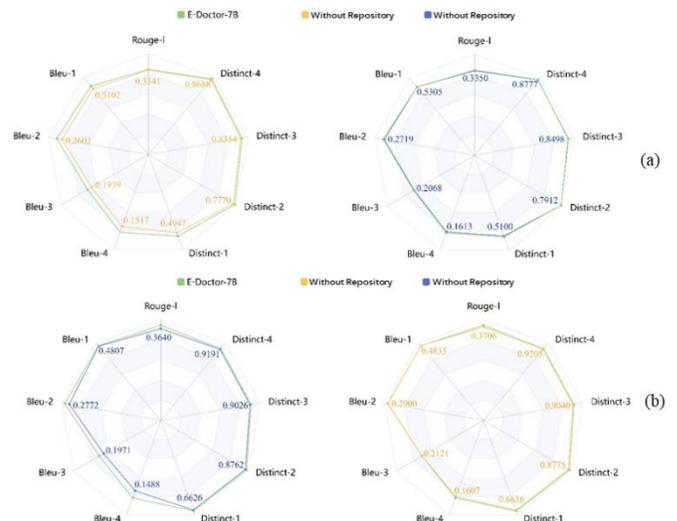

Fig. 4. (a) Single-round medical interrogation ablation experiment. (b) Multi-round medical interrogation ablation experiments

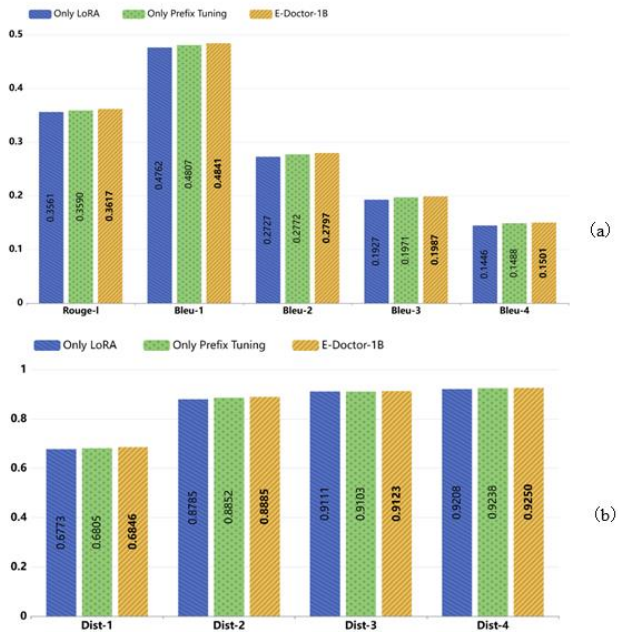

Fig. 5. (a)Graphical representation of the Rouge-l, Bleu results of the fusion-efficient parametric fine-tuning. (b)Fusion efficient parameter fine-tuning ablation experiments Distinct results illustrated.

In single-round medical consultation scenarios, the impact of knowledge source expansion is much larger than the role of doctor-patient role information enhancement. This is because the dialogue content is limited to the patient's single question and does not include the doctor's specific response information. Consequently, the patient-physician role information enhancement module lacks the support of the doctor's behavioral data, making it challenging to accurately depict the doctor's diagnostic and therapeutic patterns, thereby reducing its effectiveness. On the other hand, in the single-round medical consultation scenario, patients often describe their symptoms in detail to ensure accurate diagnosis by the doctor. This detailed description improves the matching accuracy of the exogenous knowledge base, making knowledge source expansion sufficient and effective.

In contrast, in multi-round medical consultation scenarios, the impact of knowledge source expansion is relatively lower than that of doctor-patient role information enhancement. In these scenario, the dialogue includes interaction data from both the doctor and the patient, facilitating an accurate portrayal of their behavioral characteristics through the doctor-patient role information enhancement module. This enhances the accuracy of the dialogue response. Additionally, compared with single-round medical consultation scenarios, the disclosure of patient symptoms in multi-round conversations is a gradual and in-depth process, often requiring the doctor to ask a series of targeted questions to elicit a complete presentation. At the beginning of such conversations, the lack of key disease information makes it impossible to accurately determine the user's disease type through similarity knowledge retrieval methods, resulting in less accurate and effective knowledge source expansion.

### G. Fusion of High-efficiency Parametric Fine-tuning Ablation Experiments

In the medical consultation scenario, to demonstrate the effectiveness of the hybrid efficient parameter tuning method that incorporates Prefix Tuning [11] and LoRA [12], we design an ablation experiment for the EyeDoctor-1B model in the same experimental environment for experimental validation. In the ablation experiment, "Only Prefix Tuning" indicates that only Prefix Tuning is used for parameter fine-tuning, and "Only LoRA" indicates that only LoRA is used for parameter fine-tuning.

The experimental results, presented in Fig. 9, show that both LoRA [12] and Prefix Tuning [11] contribute positively to improving dialog quality. The efficient parameter fine-tuning strategy that integrates both approaches achieves the best results across all metrics. In terms of the degree of impact, removing Prefix Tuning has a greater impact on dialog quality. This is because the doctor-patient role feature data is integrated into the large language model through Prefix Tuning parameter fine-tuning. The absence of the Prefix Tuning module results in the lack of the model's doctor-patient role perception module, directly leading to a decline in dialog quality.

## V. CONCLUSION

Ophthalmic consultation is an extremely important aspect of the medical field, requiring interaction with doctors to obtain diagnostic information for the diagnosis, treatment, and prevention of eye diseases. However, the current number of ophthalmologists is insufficient to meet the growing demand for ophthalmic consultations. By drawing an analogy between consultation dialogues and dialogue systems in natural language processing, the use of large pre-trained language models for dialogue design in specific scenarios can effectively assist consultations, inevitably enhancing efficiency and accuracy. However, using traditional fine-tuning to address Q&A tasks becomes impractical as model size and number of tasks increase. Recent studies have proposed various efficient parameter transfer learning methods, where fine-tuning a small number of additional parameters can yield significant performance improvements. In this work, we propose a multi-module fine-tuning strategy based on the differentiated styles of ophthalmic patients and doctors. By decomposing parameters, we achieved an efficient fine-tuning learning method and established a unified framework for their interconnections. Experimental results show that in ophthalmic consultation Q&A scenarios, our model EyeDoctor achieves greater accuracy in answers. Ablation experiments further demonstrate the importance of fully considering the characteristics of doctor-patient roles and exogenous knowledge augmentation for improving the algorithmic model medical interrogation.

Nonetheless, there are areas for improvement in the current model. In reality, doctors have different consultation, treatment,

and prescribing styles, which the model has not yet decomposed and modeled in such fine detail. This will be explored in depth in future work. Additionally, considering the scale of the current model, expanding the size of the baseline model could be contemplated to achieve better consultation outcomes. Furthermore, integrating multimodal data, such as medical images, videos, and patient medical history, could significantly enhance the model's diagnostic capabilities and overall effectiveness. Future research will focus on leveraging multimodal inputs to create a more comprehensive and accurate ophthalmic consultation system.

# Supplementary Materials

### SUPPLEMENTARY NOTE

In this Note, we plan to give a detailed description on Dataset. Current research in medical consultation primarily focuses on data-driven domain adaptation of generalized models using simulated data from structured medical documents or knowledge graphs. Due to the lack of real conversation data for ophthalmic disease diagnosis, this paper collects and organizes three types of data resources utilizing web crawler technology: a single-round conversation dataset from QuickAskDoctor's website, a multi-round professional conversation dataset from Ding Xiang Doctor's website, and an ophthalmology disease text knowledge base. To address non-standardized terminology and sensitive information (e.g., doctor's names, addresses) in the raw data, we pre-processed each data entry using the semantic parsing capabilities of ChatGPT. This preprocessing step removes irrelevant and sensitive information, standardizes medical terminology, and ensures the data aligns with clinical realities in ophthalmology. After that, these consultation dialogues were manually curated by clinical ophthalmologists. This approach provides a robust data foundation for developing efficient and accurate ophthalmology intelligent consultation algorithms.

### A. Single-Turn Dialogue Dataset

Quickly Ask Doctor is a Chinese online medical and health service platform created by Zhuhai Health Cloud Technology Co., Ltd. integrating the functions of disease inquiry, expert consultation and health care guidance, aiming to provide users with convenient and efficient medical and health services through the online and offline integration mode. Currently, the platform has accumulated more than 100 million single rounds of medical Q&A data, with detailed content, including Q&A content, drug recommendations, department classification, etc., to fully meet the experimental needs. In this paper, we focus on the medical Q&A data during the period of June 1 to November 10, 2022, and collect 1,292,012 single rounds of data from all departments, covering 19 first-level departments such as pediatrics, internal medicine, surgery, and 127 second-level departments. In addition to the single-round conversation data, additional information such as the type of department visited, time of visit, and recommended medications are also collected during data collection. In this paper, we will use these single-round data to perform unsupervised fine-tuning of the RoBERT a large language model to facilitate further subsequent characterization of the doctor-patient relationship and external knowledge sources.

Ophthalmology, as an important department under Pentacameral Medicine, accounted for 3.23% of the total data volume. S_Fig.2 shows a sample of a single ophthalmology consultation data collected. In response to a patient's consultation about conjunctivitis and keratitis, the doctor gave a detailed answer in terms of symptoms, diagnosis and treatment recommendations. As can be seen from the length of the dialog text, the average length of the answers given by the doctors to the questions given by the patients is more than 100 words. In the data preprocessing, we will use the powerful semantic understanding of ChatGPT [1] to further process the ophthalmology data, so as to facilitate the subsequent experimental exploration of ophthalmology as a specialized department.

### B. Multi-Turn Dialogue Dataset

Dingxiang Doctor is a famous medical and healthcare service platform with legal Internet hospital license developed by Lilac Garden team. It combines online consultation, hospital inquiry, disease self-examination, report interpretation and other services, and has more than 50,000 specialized doctors of T3A and above, which is dedicated to solving the problem of unbalanced distribution of medical resources in China. In this paper, we have collected a dataset of ophthalmology multi-round medical consultations from the website of Ding Xiang Doctor, which contains 8,647 multi-round conversations from 108 doctors. The average number of conversation rounds for a single consultation is 6.27, and the average number of conversations for a single doctor is 80.06.

S_Fig.3 shows a sample of collected multi-round conversation data from Ding Xiang Doctor ophthalmology department, in which a patient consults Dr. Huang about his mother's red and swollen eyes. Different from the generalized large language models such as ChatGPT [1] or the single-round conversations in which the doctor directly gives the user treatment suggestions, the doctor in the sample tries to get more information about the patient's symptoms in the form of questioning and finally gives the user treatment suggestions. This approach is more in line with real medical consultation scenarios, but the raw data needs to be further processed with sensitive information such as the doctor's name and the standardization of medical terminology. So we designed prompt engineering for processing and filtering these dialogues, detailed in Section D. Eventually, the processed multi-round conversation data will be applied to the following experiments to construct intelligent consultation algorithms for real ophthalmology environments. Furthermore, these consultation dialogues were manually curated by clinical ophthalmologists to ensure the highest data quality and relevance.

### C. Ophthalmic Disease Knowledge Repository

99 Health Net is an another online medical health platform, dedicated to providing users with comprehensive and detailed disease information and medical service resources. The platform covers various aspects of diseases, including overviews, symptoms, causes, diagnoses, treatments, and prevention, categorized by departments for user retrieval.

In order to improve the algorithm's ability to deal with emerging and rare diseases in the process of ophthalmic disease diagnosis [2,4], this paper collects and organizes 519 kinds of ophthalmic disease knowledge information from the 99 Disease Database website. And based on the disease knowledge information, we construct a document knowledge base, which is convenient for dynamic knowledge expansion of the algorithm. When constructing the ophthalmic disease knowledge base, 10 kinds of disease information including disease name, overview, symptoms, treatment plan, examination, identification, etiology, prevention, complications, and disease medication are collected in order. One of the single data samples is shown in S_Fig.4.

### D. Prompt Engineering Designed for multi-turn ophthalmic consultation

Prompt Engineering [3] is a specialized design technique for large language models, focusing on how to carefully design, optimize, and apply input prompts to effectively stimulate the learning ability and knowledge structure within large language models. This technique guides the model to generate output content that is more accurate and tailored to the desired objectives. Prompt Engineering is crucial for maximizing the value of large language models, as it greatly assists researchers in utilizing the models more effectively for customized natural language processing tasks.

In this study, prompt statements are designed for the original single-turn and multi-turn ophthalmic consultation dialogues. Then, leveraging the powerful semantic understanding and text generation capabilities of ChatGPT 3.5 turbo, the model is encouraged to generate a series of professionally oriented dialogues with natural fluency, appropriate length, and correct use of ophthalmic medical terminology. At the same time, privacy protection principles are also taken into consideration to reduce the risk of sensitive information leakage for doctors or patients.

The data preprocessing process based on ChatGPT [1] is illustrated in Algorithm 1. Firstly, prompt templates are set up for the original single-turn and multi-turn ophthalmic consultation dialogue data, respectively. Subsequently, the raw input content is embedded into the predetermined prompt templates to form contextually guided input data. Then, the integrated information is submitted to ChatGPT [1] for processing, resulting in output responses. Finally, upon obtaining the response output from the ChatGPT model [1], a series of rule checks are executed. Only responses that fully adhere to the preset specifications are accepted as the final standardized output results. In cases where responses do not meet the criteria, ChatGPT [1] is utilized again to generate responses, and the rule verification process is repeated.

---
**Algorithm 1: data preprocessing algorithm based on ChatGPT prompting project**

Inputs: original data set $\mathbb{R}$, corresponding hint template for the data set T, maximum number of checksums N
Output: normalized data set $\mathbb{R}'$

1) Initialization canonical data set $\mathbb{R}'$, data set not canonicalized by ChatGPT $\mathbb{R}''$, maximum number of checksums N
2) Loop over the original data set $\mathbb{R}$, for a single piece of data within the data set D
3)   Data D is embedded in the prompt template T to form real input data $D'$
4)   Loop traversal $n = 1, \dots, N$
5)     Calling the ChatGPT API to generate output O
6)     Rule checking on O
7)       If the check is successful, store the result in $\mathbb{R}'$ and end the inner loop
8)       If the check fails and $n == N$, store the check result in $\mathbb{R}''$ and end the inner loop
9)   End of loop
10) End of loop
11) Manual adjustments to the contents of the $\mathbb{R}''$ collection are merged into $\mathbb{R}'$

Output canonical data set $\mathbb{R}'$

---

The multi-turn prompt template for ophthalmic consultations is depicted in S_Fig.5(a). This template standardizes the original dialogues from eight aspects: dialogue mode, language, number of turns, content, and response format. This ensures the generation of high-quality multi-turn dialogue data. The single-turn prompt template for ophthalmic consultations is shown in S_Fig.5(b). This template standardizes the doctor-patient dialogue from four aspects, ensuring that patient inquiry information is complete and sufficient, and doctor responses are standardized and professional, thereby enhancing the usability of the data.

### SUPPLEMENTARY FIGURE

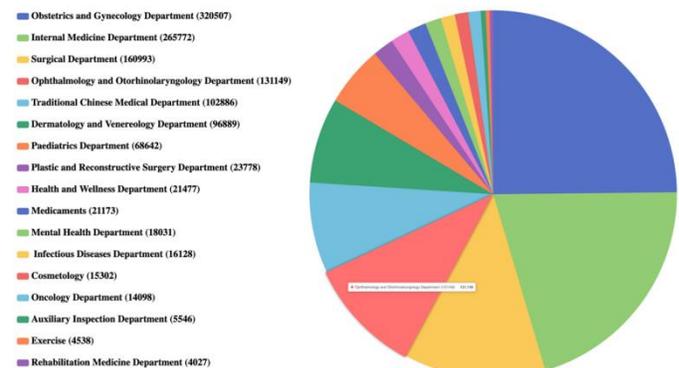

S_Fig. 1. Statistical distribution of data for single rounds of dialogues.

S_Fig. 2. Sample dialog data for a single round in ophthalmology.

S_Fig. 3. Sample Ophthalmology Multi-Round Conversation Data.

S_Fig. 4. Sample Ophthalmic Disease Knowledge Base Data.

S_Fig. 5. (a) is a template for multiple rounds of prompts for ophthalmologic consultation conversations. (b) is a template for a single round of prompts for ophthalmologic consultation conversations.